\documentclass[twoside,11pt]{article}
\usepackage{jair}

\usepackage{theapa}
\usepackage{rawfonts}
\usepackage{adjustbox}
\usepackage{amsfonts}
\usepackage{amssymb}
\usepackage{amsthm}
\usepackage{amsmath}

\usepackage{graphicx,psfrag,epsf}
\usepackage{enumerate}
\usepackage{natbib}
\usepackage{tabularx}
\usepackage{url} 

\usepackage{floatrow}
\DeclareFloatFont{small}{\small}
\floatstyle{plaintop}
\restylefloat{table}
\usepackage{soul}
\usepackage[normalem]{ulem}
\newcommand{\stkout}[1]{\ifmmode\text{\sout{\ensuremath{#1}}}\else\sout{#1}\fi}
\usepackage{comment}

\usepackage{adjustbox}
\usepackage[makeroom]{cancel}
\usepackage{bbold}
\usepackage{float}
\usepackage{bm}
\usepackage{booktabs}
\usepackage[colorlinks=true,citecolor=blue]{hyperref}%

\usepackage{algorithm}
\usepackage{algpseudocode}
\usepackage{color}

\usepackage{etoolbox}
\usepackage{csvsimple}

\definecolor{Lars}{RGB}{255,0,0}

\definecolor{Geir}{RGB}{0,0,255}

\definecolor{AH}{RGB}{255,0,255}

\definecolor{todo}{RGB}{10,200,10}

\usepackage{siunitx}

\jairheading{1}{????}{?-??}{05/23}{?/??}
\ShortHeadings{Sparsifying Bayesian neural networks with latent binary variables and normalizing flows}
{Lund-Skaaret, Storvik, Hubin}
\firstpageno{1}

\begin{document}

\title{Sparsifying Bayesian neural networks with latent binary variables and normalizing flows}

\author{\name Lars Skaaret-Lund \email lars.skaaret-lund@nmbu.no \\
       \name  Geir Storvik \email geirs@math.uio.no \\
       \name  Aliaksandr Hubin \email aliaksandr.hubin@nmbu.no \\
       }


\maketitle

\begin{abstract}
Artificial neural networks (ANNs) are powerful machine learning methods used in many modern applications such as facial recognition, machine translation, and cancer diagnostics. A common issue with ANNs is that they usually have millions or billions of trainable parameters, and therefore tend to overfit to the training data. This is especially problematic in applications where it is important to have reliable uncertainty estimates. Bayesian neural networks (BNN) can improve on this, since they incorporate parameter uncertainty. In addition, latent binary Bayesian neural networks (LBBNN) also take into account structural uncertainty by allowing the weights to be turned on or off, enabling inference in the joint space of weights and structures. In this paper, we will consider two extensions to the LBBNN method: Firstly, by using the local reparametrization trick (LRT) to sample the hidden units directly, we get a more computationally efficient algorithm. More importantly, by using normalizing flows on the variational posterior distribution of the LBBNN parameters, the network learns a more flexible variational posterior distribution than the mean field Gaussian. Experimental results show that this improves predictive power compared to the LBBNN method, while also obtaining more sparse networks. We perform two simulation studies. In the first study, we consider variable selection in a logistic regression setting, where the more flexible variational distribution leads to improved results. In the second study, we compare predictive uncertainty based on data generated from two-dimensional Gaussian distributions. Here, we argue that our Bayesian methods lead to more realistic estimates of predictive uncertainty.
\end{abstract}

\section{Introduction}\label{section1}
The idea of using a mathematical model to imitate how the brain works was first introduced in \cite{mcculloch1943logical}. However, it was not until more recent years that the true power of these models could be harnessed with the idea of using backpropagation \citep{rumelhart1986learning} to train the model with gradient descent. With the advent of modern GPU architectures, deep neural networks can be scaled to big data, and have shown to be very successful on a variety of tasks including computer vision \citep{voulodimos2018deep}, natural language processing \citep{young2018recent} and reinforcement learning \citep{li2017deep}. Modern deep learning architectures can have billions of trainable parameters \citep{khan2020survey}. Due to the large number of parameters in the model, the network has the capacity to overfit, and therefore may not generalize well to unseen data. Various regularization methods are used to try to deal with this, such as early stopping \citep{prechelt1998early}, dropout \citep{srivastava2014dropout} or data augmentation \citep{shorten2019survey}. These techniques are heuristic and therefore it is not always clear how to use them and how well they work in practice. It is also possible to reduce the number of parameters in the network with pruning. This is typically done with the dense-to-sparse method \citep{han2017dsd}. Here, a dense network is trained, while the importance of the weights (i.e. their magnitude) is recorded. Then, the weights that fall below the sparsity threshold (a hyperparameter) are removed. In \cite{frankle2018lottery}, it is hypothesized that in randomly initialized dense networks, there exists a sparse sub-network (the winning lottery ticket) that can be trained in isolation and obtain the same test accuracy as the original dense network. Instead of training and pruning once, referred to as one-shot pruning, this process is repeated sequentially several times, removing a certain percentage of the remaining weights each time, which then results in networks that have a higher degree of sparsity than the ones found with one-shot pruning. However, this comes at a higher computational cost. Further refinements to this are done in \cite{evci2020rigging}, where the network starts off dense, and dynamically removes the weights with the smallest magnitude, while also adding new connections based on gradient information. Again, these approaches are heuristic and lack a solid theoretical foundation. Another issue with deep learning models is that they often make overconfident predictions. In \cite{szegedy2013intriguing}, it was shown that adding a small amount of noise to an image can trick a classifier into making a completely wrong prediction (with high confidence), even though the image looks exactly the same to the human eye. The opposite is also possible, images that are white noise can be classified with almost complete certainty to belong to a specific class \citep{nguyen2015deep}. 

Bayesian neural networks (BNNs) were presented by~\cite{neal1992bayesian}, \cite{mackay1995bayesian}, and~\cite{bishop1997bayesian}.
They use a rigorous Bayesian methodology to handle parameter and prediction uncertainty and to incorporate prior knowledge. In many cases, this results in more reliable solutions with less overfitting; however, this comes at the expense of extremely high computational costs. Until recently, inference on Bayesian neural networks could not scale to large multivariate data due to limitations of standard Markov chain Monte Carlo (MCMC) approaches, the main quantitative procedure used for complex Bayesian inference. Recent developments of variational Bayesian approaches \citep{Gal2016Uncertainty} allow us to approximate the posterior of interest and lead to more scalable methods. 

\begin{figure}
    \centering
    \includegraphics[width=.8\textwidth]
    {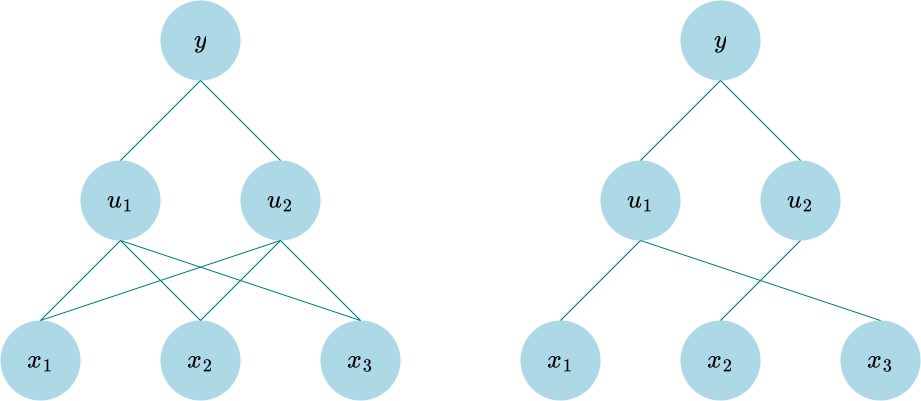}
    \caption{A dense network on the left, a possible sparse structure on the right.}
    \label{fig:nn}
\end{figure}

Still, BNNs tend to be heavily over-parameterized and difficult to interpret. It is therefore interesting to consider sparsity-inducing methods from a Bayesian perspective. This is typically done by using sparsity-inducing priors, as in variational dropout \citep{kingma2015variational,molchanov2017variational}, which uses the independent log uniform prior on the weights. This is an improper prior, meaning that it is not integrable and thus not a valid probability distribution. As noted in \cite{hron2017variational}, using this prior, combined with commonly used likelihood functions leads to an improper posterior, meaning that the obtained results can not be explained from a Bayesian modeling perspective. It is argued that variational dropout should instead be interpreted as penalized maximum likelihood estimation of the variational parameters. Additionally, \cite{gale2019state} finds that while variational dropout works well on smaller networks, it gets outperformed by the heuristic (non-Bayesian) methods on bigger networks. Another type of sparsity inducing prior is the independent scale mixture prior, where \cite{blundell2015weight} proposed a mixture of two Gaussian densities,
where using a small variance for the second mixture component leads to many of the weights having a prior around 0. Another possibility is to use the independent spike-and-slab prior, most commonly used in Bayesian linear regression models. This prior is used in the latent binary Bayesian neural networks (LBBNN) introduced by
\cite{hubin2019combining,hubin2023variational}. The prior was studied from a theoretical perspective in \citet{polson2018posterior}.  In \cite{hubin2019combining} it was empirically shown that using this prior will induce a very sparse network (around 90 $\%$ of the weights were removed) while maintaining good predictive power. The LBBNN method takes into account uncertainty around whether each weight is included or not (structural uncertainty) and uncertainty in the included weights (parameter uncertainty) given a structure, allowing for a fully Bayesian approach to network sparsification (see \hyperref[fig:nn]{Figure 1}).
In this paper, we show that transforming the variational posterior distribution with normalizing flows can result in even more sparse networks while improving predictive power compared to the mean field approach in~\cite{hubin2019combining}. Additionally, we demonstrate that the flow network handles predictive uncertainty well, and performs better than the mean-field methods at variable selection in a logistic regression setting with highly correlated variables.

\section{{The model}}
Given the explanatory variable $\boldsymbol{x} \in \mathbb{R}^n$, and the response variable $\boldsymbol{y} \in \mathbb{R}^m$, a neural network models the function
\begin{equation*}
\begin{split}
    \boldsymbol{y} &\sim f(\boldsymbol{\eta}(\boldsymbol{x})).
    \end{split}
\end{equation*}
The mean vector $\boldsymbol{\eta}$ is obtained through a composition of semi-affine transformations:
\begin{equation}\label{eq:u.act}
    u_{j}^{(l)} = \sigma^{(l)}\bigg(\sum\limits_{i=1}^{n^{(l-1)}} u_i^{(l-1)}\gamma_{ij}^{(l)}w_{ij}^{(l)} + b_j^{(l)}\bigg), j = 1,\ldots,n^{(l)}, l = 1,\ldots,L,
\end{equation}
with $\eta_{j} = u_{j}^{(L)}$. Additionally, $\bm u^{(l-1)}$ denotes the inputs from the previous layer (with $\bm u^0=\bm x$ corresponding to the explanatory variables), the $w_{ij}^{(l)}$'s are the weights, the $b_j^{(l)}$'s  are the bias terms, and $n^{(l)}$ (and $n^{(0)} = n$) the number of inputs at layer $l$ of a total $L$ layers. Further, we have the elementwise non-linear activation functions $\sigma^{(l)}$.
The additional parameters $\gamma_{ij}^{(l)}\in\{0,1\}$ denote binary inclusion variables for the corresponding weights. From here on, we drop the layer notation for readability, since the layers will always be considered independent. 

Following \cite{hubin2019combining, polson2018posterior}, we consider a structure to be defined by the configuration of the binary vector $\boldsymbol{\gamma}$, and the weights of each structure conditional on this configuration. To consider uncertainty in both structures and weights, we use the spike-and-slab prior, where for each (independent) layer of the network, we also consider the weights to be independent, resulting in the LBBNN model:
\begin{equation*}
\begin{split}
    p(w_{ij}|\gamma_{ij}) &= \gamma_{ij}\mathcal{N}(0,\sigma^{2})+(1 - \gamma_{ij})\delta(w_{ij}) \\
    p(\gamma_{ij}) &= \text{Bernoulli}(\alpha).
\end{split}
\end{equation*}
Here, $\delta(\cdot)$ is the Dirac delta function, which is considered to be zero everywhere except for a spike at zero. In addition, $\sigma^2$ and $\alpha$ denote the prior variance and the prior inclusion probability of the weights, respectively. In practice, we use the same variance and inclusion probability across all the layers and weights, but this is not strictly necessary. 

\section{Bayesian inference}
The main motivation behind using LBBNNs is that we are able to take into account both structural and parameter uncertainty, whereas standard BNNs are only concerned with parameter uncertainty. By doing inference through the posterior predictive distribution, we average over all possible structural configurations, and parameters. For a new observation $\tilde{\boldsymbol{y}}$ given data, $\mathcal{D}$, we have:
\begin{equation*}
p(\tilde{\boldsymbol{y}}|\mathcal{D})
= \sum_{\boldsymbol{\gamma}}\int_{\boldsymbol{w}}p(\tilde{\boldsymbol{y}}|\boldsymbol{w},\boldsymbol{\gamma},\mathcal{D})p(\boldsymbol{w},\boldsymbol{\gamma}|\mathcal{D})\,d\boldsymbol{w}.
\end{equation*}
This expression is intractable due to the ultra-high dimensionality of $\boldsymbol{w}$ and $\boldsymbol{\gamma}$, and using Monte Carlo sampling as an approximation is also challenging due to the difficulty of obtaining samples from the posterior distribution, $p(\boldsymbol{w},\boldsymbol{\gamma}|\mathcal{D})$. Instead of trying to sample from the true posterior, we turn it into an optimization problem, using variational inference \citep[VI,][]{blei2017variational}. The key idea is that we replace the true posterior distribution with an approximation, $q_{\boldsymbol{\theta}}(\boldsymbol{w},\boldsymbol{\gamma})$, with $\boldsymbol{\theta}$ denoting some variational parameters. We learn the variational parameters that make the approximate posterior as close as possible to the true posterior. Closeness is measured through the Kullback-Leibler (KL) divergence, 
\begin{equation*}
\text{KL} \left[q_{\boldsymbol{\theta}}(\boldsymbol{w},\boldsymbol{\gamma})||p(\boldsymbol{w},\boldsymbol{\gamma}|\mathcal{D})\right] = \sum_{\boldsymbol{\gamma}} \int_{\boldsymbol{w}} q_{\boldsymbol{\theta}}(\boldsymbol{w},\boldsymbol{\gamma})\log \dfrac{q_{\boldsymbol{\theta}}(\boldsymbol{w},\boldsymbol{\gamma})}{p(\boldsymbol{w},\boldsymbol{\gamma}|\mathcal{D})}\,d\boldsymbol{w}.
\end{equation*}
Minimizing the KL-divergence (with respect to $\boldsymbol{\theta}$) is equivalent to maximizing the evidence lower bound (ELBO):
\begin{equation} \label{eq:ELBO}
\text{ELBO}(q_{\boldsymbol{\theta}}) = \mathbb{E}_{q_{\boldsymbol{\theta}}(\boldsymbol{w},\boldsymbol{\gamma})}\left[\log p(\mathcal{D}|\boldsymbol{w},\boldsymbol{\gamma})\right] - \text{KL}\left[q_{\boldsymbol{\theta}}(\boldsymbol{w},\boldsymbol{\gamma})||p(\boldsymbol{w},\boldsymbol{\gamma})\right].
\end{equation}
The objective is thus to maximize the expected log-likelihood while penalizing with respect to the KL divergence between the prior and the variational posterior. How good the approximation becomes depends on the family of variational distributions $\{q_{\boldsymbol{\theta}},\boldsymbol{\theta}\in\Theta\}$ that is chosen.

\subsection{Choices of variational families}
A common choice \citep{blundell2015weight} for the approximate posterior in (dense) Bayesian neural networks is the mean-field Gaussian distribution. For simplicity of notation, denote now by $\textbf{W}$ the set of weights corresponding to a specific layer. Then
\begin{equation*}
q_{\boldsymbol{\theta}}({\textbf{W}})=\prod\limits_{i=1}^{n_{in}}\prod\limits_{j=1}^{n_{out}}\mathcal{N}(\tilde \mu_{ij},\tilde \sigma_{ij}^{2}),
\end{equation*}
where $n_{in}$ and $n_{out}$ denote the number of neurons in the previous and current layer, respectively. Weights corresponding to different layers are assumed independent as well.

In \cite{hubin2019combining}, the mean-field Gaussian distribution for Bayesian neural networks is extended to include the binary inclusion variables following \cite{carbonetto2012scalable}:
\begin{equation}\label{eq:modbin}
\begin{split}
q_{\boldsymbol{\theta}}({\textbf{W}|\boldsymbol{\Gamma}})&=\prod\limits_{i=1}^{n_{in}}\prod\limits_{j=1}^{n_{out}}[\gamma_{ij}\mathcal{N}(\tilde \mu_{ij},\tilde \sigma_{ij}^2) + (1-\gamma_{ij})\delta(w_{ij})]; \\
q_{\tilde \alpha_{ij}}(\gamma_{ij}) &= \text{Bernoulli}(\tilde \alpha_{ij}).
\end{split}
\end{equation}
However, the mean-field Gaussian distribution is typically too simple to be able to capture the complexity of the true posterior distribution. We follow \cite{ranganath2016hierarchical}, and introduce a set of latent variables $\boldsymbol{z}$ to model dependencies between the weights, and use the following variational posterior distribution:
\begin{equation}\label{eq:varpost}
\begin{split}
q_{\boldsymbol{\theta}}({\textbf{W}|\boldsymbol{\Gamma}},\boldsymbol{z})&=\prod\limits_{i=1}^{n_{in}}\prod\limits_{j=1}^{n_{out}}[\gamma_{ij}\mathcal{N}(z_i\tilde \mu_{ij},\tilde \sigma_{ij}^{2}) + (1-\gamma_{ij})\delta(w_{ij})]; \\
q_{\tilde \alpha_{ij}}(\gamma_{ij}) &= \text{Bernoulli}(\tilde \alpha_{ij}),
\end{split}
\end{equation}
where $\bm z$ follows a distribution $q_{\bm\phi}(\bm z)$.
\begin{figure}
    \centering
    \includegraphics[width=1.0\textwidth]{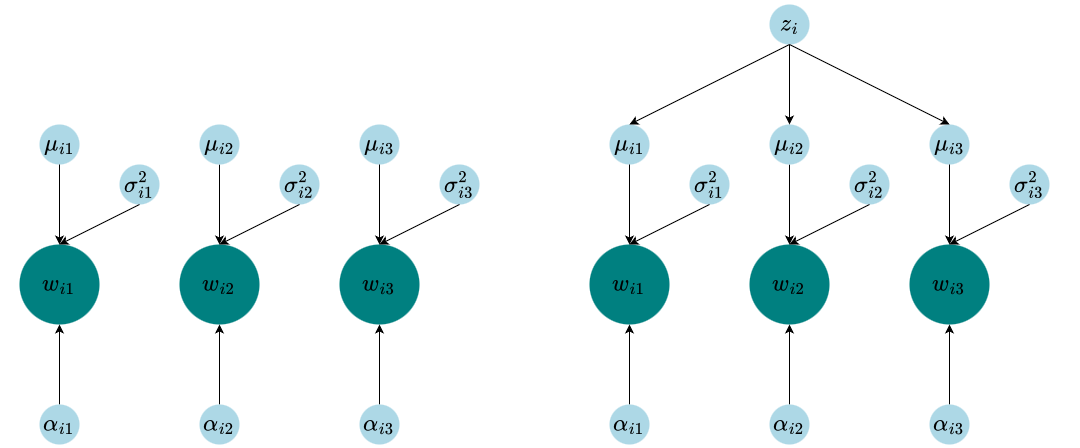}
    \caption{On the left, the mean-field variational posterior where the weights are assumed independent. On the right, the latent variational distribution $z$ allows for modeling dependencies between the weights.}
    \label{figure:hierarchicalvi}
\end{figure}
For an illustration of the difference between the two variational distributions~\eqref{eq:modbin} and~\eqref{eq:varpost}, see \hyperref[figure:hierarchicalvi]{Figure 2}. Our novel variational distribution is thus able to take into account both weight and structural uncertainty, in addition to modeling dependencies between the weights.  As for $\textbf{W}$, also $\bm z$ is a set of variables related to a specific layer, independence between layers is assumed also for $\bm z$'s. To increase the flexibility of the variational posterior, we apply normalizing flows \citep{rezende2015variational} to $q_{\boldsymbol{\phi}}(\boldsymbol{z})$. In general, a normalizing flow is a composition of invertible transformations of some initial (simple) random variable $\textbf{z}_0$,
\begin{equation*}
\bm z_k=f_k(\bm z_{k-1}),\quad k=1,...,K.
\end{equation*}
The log density of the transformed variable $\bm z=\boldsymbol{z}_K$ is given as,
\begin{equation}\label{eq:logdet}
\log q_K(\boldsymbol{z}_K) = \log q_0(\boldsymbol{z}_0) - \sum \limits_{k=1}^{K} \log \left|\det \frac{\partial\bm z_k}{\partial \bm z_{k-1}} \right|.
\end{equation}
We are typically interested in transformations that have a Jacobian determinant that is tractable, and fast to compute, in addition to being highly flexible. Transforming the variational posterior distribution in a BNN with normalizing flows was first done in \cite{louizos2017multiplicative}, who coined the term multiplicative normalizing flows (MNF), where the transformations were applied in the activation space instead of the weight space. As the weights are of much higher dimensions, the number of flow parameters and thus the number of parameters of variational distribution would explode quickly. We will do the same here. The main difference in our work is that by using the variational posterior in \eqref{eq:varpost}, we also get sparse networks.  

For the normalizing flows, we will use the inverse autoregressive flow (IAF), with numerically stable updates, introduced by \cite{kingma2016improved}. It works by transforming the input in the following way:
\begin{equation}\label{eq:IAF}
\begin{split}
\boldsymbol{z}_{k-1} &= \text{input} \\
\boldsymbol{m}_k,\boldsymbol{s}_k &= \text{NeuralNetwork}(\boldsymbol{z}_k) \\ 
\boldsymbol{\kappa}_k &= \text{sigmoid}(\boldsymbol{s}_k) \\
\boldsymbol{z}_{k} &= \boldsymbol{\kappa}_k \odot \boldsymbol{z}_{k-1} + (1 - \boldsymbol{\kappa}_k) \odot \boldsymbol{m}_k,
\end{split}
\end{equation}
where $\odot$ denotes elementwise multiplication. 
Assuming the neural network in \eqref{eq:IAF} is autoregressive 
(i.e $z_{k,i}$ can only depend on $z_{k,1:i-1}$),
we get a lower triangular Jacobian and
\begin{equation}
    \log \left |\det\dfrac{\partial \boldsymbol{z}_{k}}{\partial \boldsymbol{z}_{k-1}} \right| = \sum\limits_{i=1}^{n_{in}}\log \kappa_{k,i}.
\end{equation}

\subsection{Computing the variational bounds}
In practice, we minimize the negative ELBO in \eqref{eq:ELBO}. In order to compute this upper bound, we need to marginalize out $\boldsymbol{z}$ from the joint posterior distribution (still within one layer, dropping the layer notation and also from here on dropping the subscript of $q(\cdot)'s$ for variational parameters for simplified notation):
\begin{equation*}
   q(\boldsymbol{W,\Gamma}) = \int q(\boldsymbol{W,\Gamma},\boldsymbol{z}) \,d\boldsymbol{z}.
\end{equation*}
This expression is generally not tractable, therefore we must turn to an approximation to learn its parameters. Similarly to \cite{louizos2017multiplicative}, we use that
\begin{equation*}
    \log q(\boldsymbol{W,\Gamma})= \log q(\boldsymbol{W,\Gamma}|\boldsymbol{z})+ \log q(\boldsymbol{z}) - \log q(\boldsymbol{z}|\boldsymbol{W,\Gamma}).
\end{equation*}
 We thus get the following expression for the KL-divergence,
\begin{equation}\label{eq:bound12}
\begin{split}
\text{KL}&\left[q(\boldsymbol{W,\Gamma})||p(\boldsymbol{W,\Gamma})\right]=\\
&\mathbb{E}_{q(\boldsymbol{W,\Gamma},\boldsymbol{z})}\biggl[\text{KL}\left[q(\boldsymbol{W,\Gamma}|\boldsymbol{z})||p(\boldsymbol{W,\Gamma})\right]
+\log q(\boldsymbol{z})-\log q(\boldsymbol{z}|\boldsymbol{W,\Gamma})\Biggl].
\end{split}
\end{equation}
After doing some algebra, we get the following for the first term:
\begin{align*}
\text{KL}&\left[q(\boldsymbol{W,\Gamma}|\boldsymbol{z})||p(\boldsymbol{W,\Gamma})\right] \\
    =& \sum \limits_{ij}\left[ \tilde \alpha_{ij}\Biggl(\log\tfrac{\sigma_{ij}}{\tilde \sigma_{ij}}+\log\tfrac{\tilde \alpha_{ij}}{\alpha_{ij}}
    -\frac{1}{2}+\tfrac{\tilde \sigma_{ij}^2+(\tilde \mu_{ij}z_i-\mu_{ij})^2}{2\sigma_{ij}^2}\Biggl)+
 (1-\tilde \alpha_{ij})\log \tfrac{1-\tilde \alpha_{ij}}{1-\alpha_{ij}}\right].
\end{align*}
The second term is simply
\begin{equation*}
\log q_K(\boldsymbol{z}) = \log q_0(\boldsymbol{z}_0) - \sum\limits_{i=1}^{n_{in}}\log \kappa_{k,i}.
\end{equation*}
The third term, $q(\boldsymbol{z}|\boldsymbol{W,\Gamma})$, is in general, intractable and also difficult to compute numerically. To address this, we introduce an additional auxiliary distribution $r_{\theta}(\boldsymbol{z}|\boldsymbol{W,\Gamma})$, parameterized by $\theta$ and get the  upper bound of \eqref{eq:bound12} following \citet{ranganath2016hierarchical}.
\begin{equation}\label{eq:bound2}
\begin{split}
\text{KL}&\left[q(\boldsymbol{W,\Gamma})||p(\boldsymbol{W,\Gamma})\right]\leq\\
&\mathbb{E}_{q(\boldsymbol{W,\Gamma},\boldsymbol{z})}\bigg[\text{KL}\left[q(\boldsymbol{W,\Gamma}|\boldsymbol{z})||p(\boldsymbol{W,\Gamma})\right]
+\log q(\boldsymbol{z})-\log r(\boldsymbol{z}|\boldsymbol{W,\Gamma})\bigg].
\end{split}
\end{equation}
This bound is looser than the original upper bound (see \cite{ranganath2016hierarchical} for a proof), but the dependence structure in the variational posterior distribution can compensate for this. 
For the last term in \eqref{eq:bound2}, $\log r(\boldsymbol{z}|\boldsymbol{W,\Gamma})$, we follow \cite{louizos2017multiplicative} and use inverse normalizing flows to make this distribution flexible, with
\begin{equation*}
    r_B(\boldsymbol{z}_B|\boldsymbol{W,\Gamma}) =\prod \limits_{i =1}^{n_{in}} \mathcal{N}(\nu_i,\tau_i^2).
\end{equation*}
We define the dependence on $\boldsymbol{W}$ and $\boldsymbol{\Gamma}$ similar to \cite{louizos2017multiplicative}:

\begin{equation}
\begin{split}
    \boldsymbol{\nu} &=n_{\text{out}}^{-1} (\bm d_1\bm s^T)\bm 1,\hspace*{2cm}\text{with } \bm s=\sigma(\textbf{e}^T (\boldsymbol{W}\odot\boldsymbol{\Gamma}))\\
     \log \boldsymbol{\tau^2}&=n_{\text{out}}^{-1}(\bm d_2\bm s^T)\bm 1.
\end{split}
\end{equation}
Here, $\textbf{d}_1$, $\textbf{d}_2$ and $\textbf{e}$ are trainable parameters with the same shape as $\boldsymbol{z}$. For $\sigma$, we use \href{https://pytorch.org/docs/stable/generated/torch.nn.Hardtanh.html}{hard-tanh}, as opposed to tanh (used in \cite{louizos2017multiplicative}) as this works better empirically. For the last term of~\eqref{eq:bound2}, we thus have:
\begin{equation*}
    \log r\left(\boldsymbol{z}|\textbf{W},\boldsymbol{\Gamma}\right) = \log r_B\left(\boldsymbol{z}_B|\textbf{W},\boldsymbol{\Gamma}\right) +  \log \left|\det \frac{\partial \boldsymbol{z}_B}{\partial \boldsymbol{z}} \right|.   
\end{equation*}
This means that we must use two normalizing flows,  one to get from $\boldsymbol{z}_0$ to $\bm z=\boldsymbol{z}_K$, and another from $\boldsymbol{z}_B$ to $\boldsymbol{z}$. Here, we have shown the inverse normalizing flow with only one layer, but can in general be extended to an arbitrary number of them just like in \eqref{eq:logdet}. 

For the biases, we assume they are independent of the weights, and each other. We use the standard normal prior with the mean-field Gaussian approximate posterior. As we do not use normalizing flows on the biases, we only need to compute the KL-divergence between two Gaussian distributions:
\begin{equation*}
\text{KL}\left[q(\boldsymbol{b})||p(\boldsymbol{b})\right]
=\sum\limits_{ij}\left[
   \log\frac{\sigma_{b_{ij}}}{\tilde \sigma_{b_{ij}}}-
        \frac{1}{2}+
        \frac{\tilde \sigma_{b_{ij}}^2+(\tilde \mu_{b_{ij}}-\mu_{b_{ij}})^2}{2\sigma_{b_{ij}}^2}\right].
\end{equation*}
In practice, the ELBO is optimized through a (stochastic) gradient algorithm where the reparametrization trick \citep{kingma2013auto} combined with mini-batch is applied. 

\section{Extending LBNNNs with the LRT and MNF}
One downside to the LBBNN method, proposed by \cite{hubin2019combining,hubin2023variational}, is that each forward pass during training requires sampling of the large $\boldsymbol \Gamma$ and $\textbf{W}$ matrices, consisting of all $\gamma_{ij}$'s, and $w_{ij}$'s, to compute the activations for each layer in the network for the stochastic variational inference procedure described in detail in \cite{hubin2019combining,hubin2023variational}. Additionally, due to the binary nature of the $\gamma_{ij}$'s, a continuous relaxation was required in \cite{hubin2019combining,hubin2023variational}. Here, we will show how to circumvent both of these issues by sampling the pre-activations $h_j$ given in~\eqref{eq:u.act} directly, typically referred to as the local reparametrization trick (LRT) \citep{kingma2015variational}. The difference in our case is that we must also take into account the binary inclusion variables. Here when we refer to $h_j$, we shall mean the activation before the non-linear activation function is applied. Then, we still use exactly the same stochastic variational inference optimization algorithm as in \cite{hubin2019combining,hubin2023variational} except for not requiring any relaxations. We can compute the mean and the variance of this as:
\begin{equation*}
\begin{split}
\mathbb{E}(h_j)&=\mathbb{E}\left[b_j+\sum\limits_{i=1}^{N
} [o_i\gamma_{ij}w_{ij}] \right]= \tilde \mu_{b_{j}}+\sum\limits_{i=1}^N [o_{i}\tilde \alpha_{ij}\tilde \mu_{ij} ] \\
\text{Var}(h_j)&=\text{Var}\left[b_j +\sum\limits_{i=1}^{N} [o_i\gamma_{ij}w_{ij}] \right]= \tilde \sigma_{b_{j}}^2 +\sum\limits_{i=1}^N [o_i^2\tilde \alpha_{ij}(\tilde \sigma_{ij}^2 + (1-\tilde \alpha_{ij})\tilde \mu_{ij}^2)].
\end{split} 
\end{equation*}
Here, $o$ denotes the output from the previous layer, consisting of $N$ neurons. The general idea behind the LRT is that if we have a sum of independent Gaussian random variables, the sum will also be (exactly) Gaussian. In our case, we have a mixture of independent Gaussians, but the central limit theorem still holds for a sum of independent random variables, as long as Lindeberg's condition is satisfied. We also verify empirically that a sample of activations generated using the LRT will follow approximately the same distribution as a sample of activations generated by sampling $\boldsymbol \Gamma$ and $\textbf{W}$. We can thus sample the activations as (independent) Gaussian variables with the means and variances given from the formulas above. 
Also, if we use the LRT, we have a reduction in the variance of the gradient estimates, as shown in \cite{kingma2015variational}. Note also that the approximations induced by the sampling procedure for $\bm h$ also can be considered as an alternative variational approximation directly for $p(\bm h|\mathcal{D})$.

For our second extension, we apply normalizing flows in the activation space to increase the flexibility of the variational posterior. When using normalizing flows, the mean and the variance of the activation $h_j$ are:
\begin{equation*}
\begin{split}
\mathbb{E}(h_j)&=\sum\limits_{i=1}^N o_{i}z_{i}\tilde \alpha_{ij}\tilde \mu_{ij}  + \tilde \mu_{b_{j}} \\
\text{Var}(h_j)&=\sum\limits_{i=1}^N o_i^2\tilde \alpha_{ij}(\tilde \sigma_{ij}^2 + (1-\tilde \alpha_{ij})z_{i}^2\tilde \mu_{ij}^2) + \tilde \sigma_{b_{j}}^2,
\end{split} 
\end{equation*}
It should be noted that $\boldsymbol{z}$ affects both the mean and the variance of our Gaussian approximation, whereas in \cite{louizos2017multiplicative} it only influences the mean. \cite{louizos2017multiplicative} also sample one $\boldsymbol{z}$ for observation within the mini-batch.
We found that empirically it made no difference on performance to only sample one vector and multiply the same $\boldsymbol{z}$ with each input vector. We do this, as it is more computationally efficient.

\section{Experiments}
\subsection{Background}
In this section, we are mainly interested in comparing the baseline method of \cite{hubin2023variational}, (denoted LBBNN-GP-MF in their paper), denoted LBBNN here, with the two extensions detailed in this paper. The first one, where we use the local reparametrization trick (LRT), we denote LBBNN-LRT, whereas the second one, where we use multiplicative normalizing flows (and the LRT) on the variational posterior distribution we shall denote LBBNN-FLOW. In \cite{hubin2019combining, hubin2023variational}, comprehensive classification experiments show that LBBNNs can sparsify Bayesian neural networks to a large degree while maintaining high predictive power. Additionally, \cite{hubin2023variational} consider an experiment where dependencies in the variational posterior distribution are built-in by using a multivariate normal distribution on the logit scale for the posterior inclusion parameters. The multivariate normal option does not improve performance compared to the mean-field  approach in the examples considered in~\citet{hubin2023variational}, hence we drop this option from our set of compared approaches.
Here, we demonstrate that increasing the flexibility of the variational posterior with normalizing flows improves predictive power compared to the approaches considered in \cite{hubin2019combining,hubin2023variational} for a set of baseline classification problems. Additionally, we perform two simulation studies. In the first one, we consider variable selection in a logistic regression setting, with highly correlated explanatory variables. In the second, we generate data from clusters of two-dimensional Gaussian distributions and compare how the different methods handle predictive uncertainty. All the experiments were coded in Python, using the PyTorch deep learning library \citep{NEURIPS2019_9015}.

\subsection{Classification experiments}
We perform two classification experiments, one with a fully connected architecture~\citep[as in ][]{hubin2019combining,hubin2023variational}, and the other with a convolutional architecture (see appendix \ref{appendix:A} for details on how this is implemented). In both cases, we classify on MNIST \citep{deng2012mnist}, FMNIST (Fashion MNIST) \citep{xiao2017fashion} and KMNIST (Kuzushiji MNIST) \citep{clanuwat2018deep}. MNIST is a database of handwritten digits ranging from 0 to 9. FMNIST consists of ten different fashion items from the Zalando (Europe's largest online fashion retailer) database. Lastly, KMNIST also consists of ten classes, with each one representing one row of Hiragana, a Japanese syllabary. All of these datasets contain 28x28 grayscale images, divided into a training and validation set with 60\,000 and 10\,000 images respectively. MNIST and FMNIST are well-known and often utilized datasets, so it is easy to compare performance when testing novel algorithms. KMNIST is a somewhat recent addition and is considered a more challenging task than the classical MNIST digits dataset because each Hiragana can have many different symbols.

\begin{table}[!h]
\begin{tabular}{@{}lccccccccc@{}}

\toprule
           \textbf{KMNIST }& \multicolumn{4}{c}{Median probability model} & \multicolumn{4}{c}{Full model averaging} & \\ \midrule
Method     & min     & median   & max & density    & min       & median     & max       & density\\
LBBNN      & 89.22   & 89.59    & 89.98  & 0.113  & 89.43     & 89.76      & 90.21     & 1.000 \\
LBBNN-LRT  & 90.04   & 90.26    & 90.43 & 0.136  & 90.23     & 90.39      & 90.60    & 1.000 \\
LBBNN-FLOW & 90.64   & \textbf{91.12}    & 91.46 & 0.096   & 91.16     & \textbf{91.30}      & 91.61    & 1.000   \\ \bottomrule
\toprule
         \textbf{MNIST} & \multicolumn{4}{c}{Median probability model} & \multicolumn{4}{c}{Full model averaging} &         \\ \midrule
Method     & min     & median   & max & density    & min       & median     & max       & density \\
LBBNN      & 98.01   & 98.10    & 98.20 & 0.098  & 98.03     & 98.14      & 98.23     & 1.000   \\
LBBNN-LRT  & 97.84   & 97.95    & 98.09 & 0.103   & 98.01     & 98.08      & 98.11     & 1.000   \\
LBBNN-FLOW & 98.14   & \textbf{98.36}    & 99.42 & 0.074   & 98.23     & \textbf{98.42}      & 98.53     & 1.000   \\ \bottomrule
\toprule
          \textbf{FMNIST} & \multicolumn{4}{c}{Median probability model} & \multicolumn{4}{c}{Full model averaging} &         \\ \midrule
Method     & min     & median   & max  & density   & min       & median     & max       & density \\
LBBNN      & 88.47   & 88.76    & 88.90 & 0.106   & 88.60     & 88.74      & 88.91     & 1.000   \\
LBBNN-LRT  & 87.51   & 87.82    & 87.94 & 0.141   & 87.88     & 87.94      & 88.14     & 1.000   \\
LBBNN-FLOW & 89.49   & \textbf{89.70}    & 89.88 & 0.097   & 89.52     & \textbf{89.80}      & 89.92     & 1.000   \\ \bottomrule
\end{tabular}
\caption{Performance metrics on the KMNIST, MNIST, FMNIST validation data, for the fully connected architecture. For the accuracies (\%), we report the minimum, maximum, and median over the ten different runs. Density is computed as an average over the ten runs.}
\label{table:fcbnn_metrics}
\end{table}

For the experiments with the fully connected architecture, we use the same set-up as in \cite{hubin2019combining,hubin2023variational}. We have two hidden layers with 400 and 600 neurons respectively, ReLU \citep{agarap2018deep} activation functions, and the Adam \citep{kingma2014adam} optimizer. We use a batch size of 100 and train for 250 epochs. All the experiments are run 10 times, and we report the minimum, median, and maximum predictive accuracy over these 10 runs. The reported density (1-sparsity) is an average over these 10 runs. For the LBBNN-LRT and LBBNN-FLOW methods, we use the standard normal prior for all the weights and biases in the network, and a prior inclusion probability of 0.10. For both $q(\boldsymbol{z})$ and $r(\boldsymbol{z}|\textbf{W},\boldsymbol{\Gamma})$, we use flows of length two, where the neural networks consist of two hidden layers with 250 neurons each. For our second classification experiment, we use the LeNet-5 \citep{lecun1998gradient} convolutional architecture, but with 32 and 48 filters for the convolutional layers. We use the same priors and normalizing flows as in the previous experiment, and the same datasets.  

To measure predictive performance, we consider two approaches. First, the fully Bayesian model averaging approach, where we average over 100 samples from the variational posterior distribution, taking into account uncertainty in both weights and structures. Secondly, we consider the median probability model \citep{barbieri2004optimal}, where we only do model averaging over the weights that have a posterior inclusion probability greater than 0.5 following \cite{hubin2019combining,hubin2023variational}, whilst others are excluded from the model. This allows for significant sparsification of the network. We emphasize that this is possible because we can go back to sampling the weights when doing inference. We also report the density, i.e. the proportion of weights included in the median probability model.

\begin{table}[!h]
\begin{tabular}{@{}lccccccccc@{}}
\toprule
          \textbf{KMNIST} & \multicolumn{4}{c}{Median probability model} & \multicolumn{4}{c}{Full model averaging} & \\ \midrule
Method     & min     & median   & max & density    & min       & median     & max       & density\\
LBBNN      & 95.13   & 95.52    & 95.89  & 0.359  & 95.21     & 95.48      & 95.78     & 1.000 \\
LBBNN-LRT  & 94.73   & 94.94    & 95.16 & 0.429  & 95.07     & 95.42      & 95.65    & 1.000 \\
LBBNN-FLOW & 95.73   & \textbf{95.99}    & 96.43 & 0.351   & 96.00     & \textbf{96.18}      & 96.44    & 1.000   \\ \bottomrule
\toprule
          \textbf{MNIST} & \multicolumn{4}{c}{Median probability model} & \multicolumn{4}{c}{Full model averaging} &         \\ \midrule
Method     & min     & median   & max & density    & min       & median     & max       & density \\
LBBNN      & 99.22   & 99.26    & 99.35 & 0.353  & 99.21     & 99.28      & 99.33     & 1.000   \\
LBBNN-LRT  & 99.11   & 99.26    & 99.31 & 0.406   & 99.20     & 99.28      & 99.34     & 1.000   \\
LBBNN-FLOW & 99.15   & \textbf{99.27}    & 99.41 & 0.338   & 99.16     & \textbf{99.29}      & 99.42     & 1.000   \\ \bottomrule
\toprule
          \textbf{FMNIST} & \multicolumn{4}{c}{Median probability model} & \multicolumn{4}{c}{Full model averaging} &         \\ \midrule
Method     & min     & median   & max  & density   & min       & median     & max       & density \\
LBBNN      & 91.14   & 91.31    & 91.48 & 0.352   & 91.10     & 91.26      & 91.44     & 1.000   \\
LBBNN-LRT  & 90.04   & 90.40    & 90.85 & 0.433   & 90.52     & 90.73      & 91.06     & 1.000   \\
LBBNN-FLOW & 90.52   & \textbf{91.54}    & 91.75 & 0.367   & 91.38     & \textbf{91.71}      & 92.04     & 1.000   \\ \bottomrule
\end{tabular}
\caption{Performance metrics on the KMNIST, MNIST, FMNIST validation data, with the convolutional architecture. See the caption in Table~\ref{table:fcbnn_metrics} for more details.}
\label{table:cnn_metrics}
\end{table}

The results with the fully connected architecture can be found in Table~\ref{table:fcbnn_metrics} and for the convolutional architecture in Table~\ref{table:cnn_metrics}. Firstly, we see that using the LBBNN-LRT gives results that are comparable to the baseline LBBNN method, except for FMNIST where it performs a bit worse both with the fully connected and with the convolutional architecture. It is no surprise that these results are similar, as using the LRT is mainly a computational advantage. Secondly, we note that the LBBNN-FLOW method performs better than the other two methods, on both convolutional and fully connected architectures, while having the most sparse networks. We note that our method's accuracy on MNIST with the convolutional architecture is comparable to the accuracy reported in \cite{louizos2017multiplicative} ($99.30\%)$. This then suggests that it is possible to sparsify convolutional BNNs without losing much predictive power. The higher density in general on the convolutional architectures is mainly a result of slightly different parameter initializations, however, these networks could also be sparsified to a similar degree as the fully connected ones. The increased predictive power of using normalizing flows comes at a computational cost. With the fully connected architecture, we observed that it took around 4 seconds to train one epoch with LBBNN-LRT, 13 seconds with LBBNN, and 17 seconds with LBBNN-FLOW on an NVIDIA A10 GPU. On the convolutional architecture, it took 7 seconds per epoch with the LBBNN-LRT, 18 seconds with LBBNN, and 28 with normalizing flows.

\subsection{Logistic regression simulation study}
In this section, we do a variable selection experiment within a logistic regression setting. As logistic regression is just a special case of a neural network with one neuron (and hence one layer), modifying the algorithms is straightforward. We are limiting ourselves to the logistic regression context to be able to compare to the original baseline method from \cite{carbonetto2012scalable}, who have shown that the mean-field variational approximation starts to fail the variable selection task when the covariates are correlated. We use the same data as in \cite{2018}, consisting of a mix of 20 binary and continuous variables, with a binary outcome, and we have 2\,000 observations. The covariates, $\boldsymbol{x}$, are generated with a strong and complicated correlation structure between many of the variables (see \hyperref[fig:corr]{Figure 3}). For more details on exactly how the covariates are generated, see appendix B of \cite{2018}. The response variable, $y$, is generated according to the following data-generating process:
\begin{equation*}
\begin{split}
\eta &\sim \mathcal{N}(\boldsymbol{\beta}\boldsymbol{x},0.5)\\
y &\sim \text{Bernoulli}\left(\frac{\exp({\eta})}
{1 + \exp({\eta}) }\right)
\end{split}
\end{equation*}
with the regression parameters defined to be:
\begin{equation*}
\boldsymbol{\beta} = (-4, 0, 1, 0, 0, 0, 1, 0, 0, 0, 1.2, 0, 37.1, 0, 0, 50,- 0.00005, 10, 3, 0).
\end{equation*}
\begin{figure}
    \centering
    \includegraphics[width=0.8\textwidth]{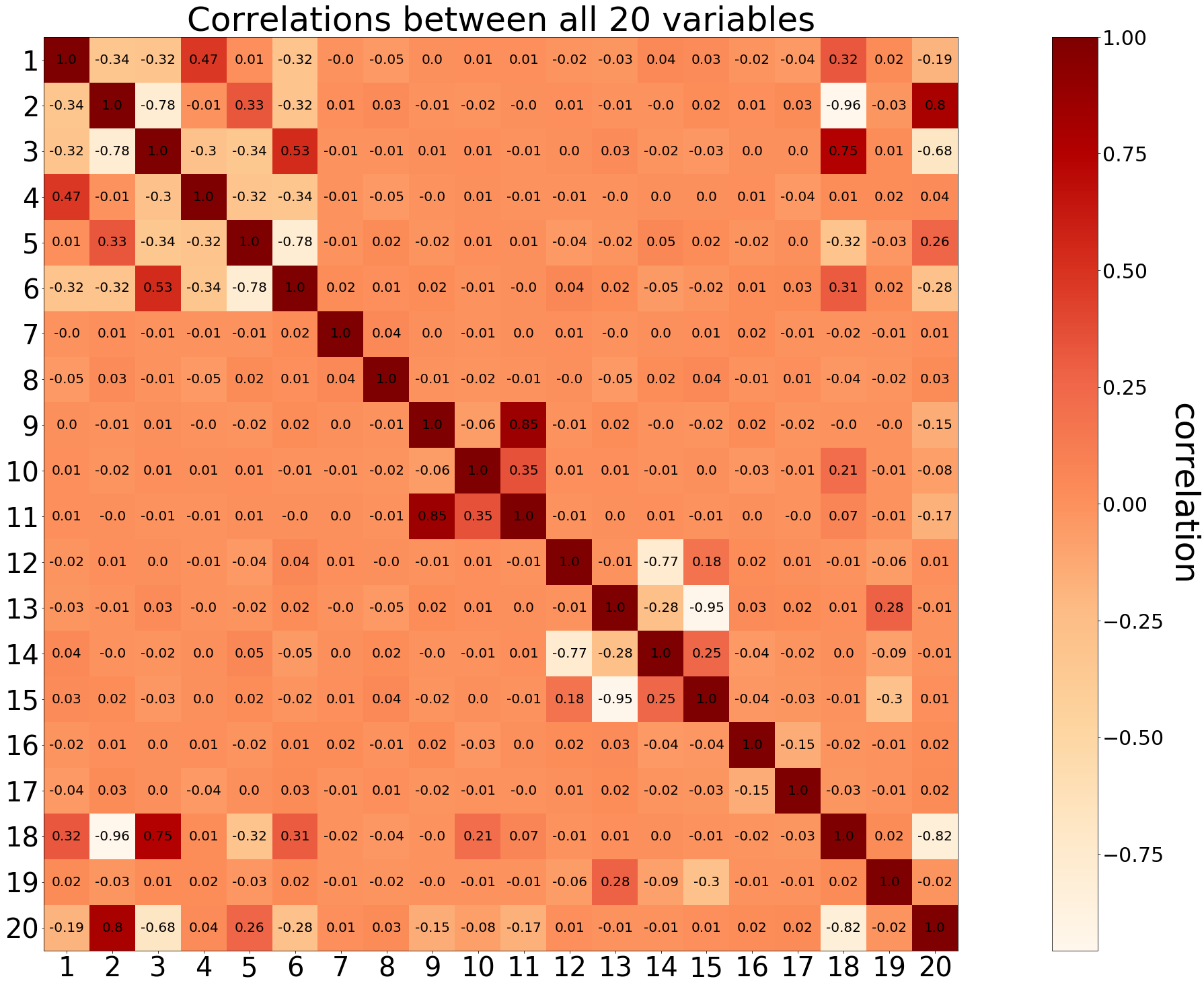}
    \caption{Plots showing the correlation between different variables in the logistic regression simulation study.}\label{fig:corr}
\end{figure}

The goal is to train the different methods to select the non-zero elements of $\boldsymbol{\beta}$. We consider  the parameter $\beta_j$ to be included if the posterior inclusion probability $\alpha_j > 0.5$, i.e. the median probability model of \cite{barbieri2004optimal}. We fit the different methods 100 times (to the same data), each time computing the true positive rate (TPR), and the false positive rate (FPR):
\begin{equation*}
\begin{split}
    \text{TPR} &= \sum\limits_{j = 1}^M\frac{\text{TP}_j}{\text{TP}_j+\text{FN}_j},\\
     \text{FPR} &= \sum\limits_{j = 1}^M\frac{\text{FP}_j}{\text{FP}_j+\text{TN}_j}.
\end{split}
\end{equation*}
Here $n = 20$ variables, and \text{TP} = true positive, meaning that a non-zero weight was correctly included. \text{FN} = false negative, meaning a non-zero weight was not included.  \text{FP} = false positive, meaning a weight that was zero was included. \text{TN} = true negative, meaning that a zero weight was not included. Thus, \text{TPR} measures the proportion of variables with non-zero weights correctly included, whereas \text{FPR} measures the proportion of variables with zero weights that were wrongly included. 

\begin{table}[t]
\begin{tabular}{@{}ccccc@{}}
\toprule
         & CS &  LBBNN-LRT & LBBNN-FLOW    \\ \midrule
mean TPR & 0.681   & 0.838      & \textbf{0.972} \\
mean FPR & 0.125   & 0.084      & \textbf{0.074 }\\ \bottomrule
\end{tabular}
\caption{Performance metrics on the logistic regression variable selection simulation study.}
\label{table:simstudy}
\end{table}

In this experiment we compare our approaches LBBNN-LRT and LBBNN-FLOW against  the algorithm proposed by \cite{carbonetto2012scalable}, denoted as CS henceforth. That method is very similar to LBBNN-LRT, as it uses the same variational distribution, but optimization is done with importance sampling, and coordinate ascent variational inference (without subsampling from the data). For the normalizing flows, we use flows of length two with the neural networks having two hidden layers of 100 neurons each. We use a batch size of 400 and train for 500 epochs. We use standard normal priors for the weights and a prior inclusion probability of 0.25 on the inclusion indicators for all three approaches. Hence we are in the setting of a Bayesian logistic regression, with variable selection. 

\begin{figure}[!h]
    \centering
    \includegraphics[width=0.325\textwidth]{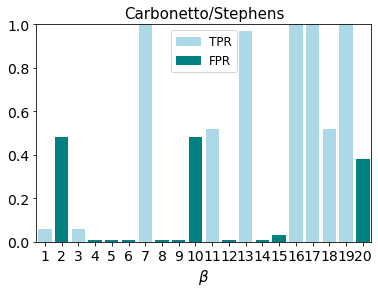}
    \includegraphics[width=0.325\textwidth]{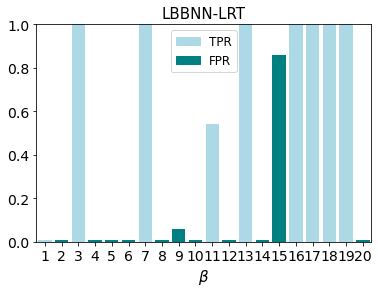}
    \includegraphics[width=0.325\textwidth]{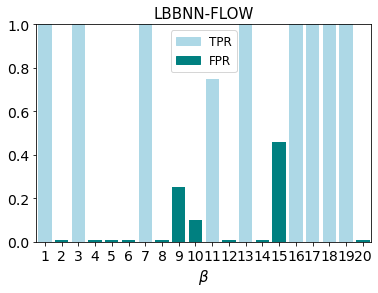}
    \caption{Bar-plots showing how often the weights are included over 100 runs.}\label{fig:barplot}
\end{figure}

The results are in \hyperref[table:simstudy]{Table 3}. We also show a bar-plot (\hyperref[fig:barplot]{Figure 4}) for each of the 20 weights over the 100 runs. We see that LBBNN-FLOW performs best, with the highest TPR and the lowest FPR. It is especially good at picking out the correct variables where there is a high correlation between many of them (for example $\beta_1 - \beta_6$). We might attribute this to the more flexible variational posterior distribution, as opposed to the mean-field Gaussian distribution used in the other three methods. \cite{carbonetto2012scalable} also discuss how the mean-field approach can only be expected to be a good approximation when the variables are independent or at most weakly correlated.

\subsection{Predictive uncertainty}

A key motivation behind using BNNs is their ability to handle predictive uncertainty more accurately than non-Bayesian neural networks. We therefore in this experiment want to illustrate how our approaches LBBNN-LRT and LBBNN-FLOW, as well as Monte Carlo dropout \citep{gal2016dropout}, and a regular (dense) BNN behave in terms of the predictive uncertainty. The purpose of this study is, thus, illustrative rather than comparative and the methods are not competing here. For this experiment, we simulate 5 clusters of data from two-dimensional Gaussian distributions. For the five Gaussians, we use the means and covariances reported in Appendix \ref{appendix:B}.
The data is then transformed to be in the range between 0 and 1, for ease of visualization. The task is to classify to the correct class corresponding to a specific cluster.\begin{figure}[!h]
    \centering
    \includegraphics[width=0.47\textwidth]
    {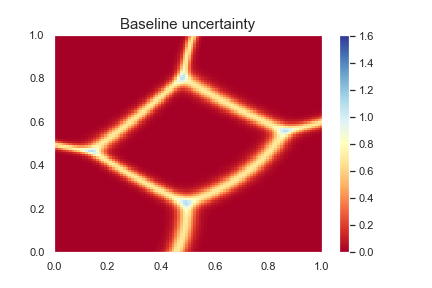}
    \caption{The data generative uncertainty of the Gaussian clusters in the predictive uncertainty experiment.}
    \label{figure:baseuncert}
\end{figure} We generate three datasets, with 10, 50, and 200 samples from each class, respectively. For all the methods, we fit a network with one hidden layer consisting of 1000 neurons, meaning we are in a setting where the number of trainable parameters is much larger than the number of observations, which is a typical scenario for applications of BNN. For dropout, we use 0.5 for the dropout probability, and we use 0.5 for the prior inclusion probabilities for LBBNN-LRT and LBBNN-FLOW. We use flows of length two, with the neural networks consisting of two hidden layers of 50 neurons each. For all the methods, we use 10 samples for model averaging. To measure predictive uncertainty, we generate a test set over a grid over $[0,1]^2$ and compute the entropy of the predictive distributions for each point in the grid. Maximum entropy is attained when the predictive distribution is uniform, i.e. 0.2 for each class. The results are shown in \hyperref[figure:entropy50]{Figure 6}, \hyperref[figure:entropy250]{Figure 7}, and \hyperref[figure:entropy1000]{Figure 8}. In addition, we show the entropy of the data generative model in \hyperref[figure:baseuncert]{Figure 5}, to which the predictive entropies are expected to converge as the sample size increases. 

With little data, we see a stark difference between dropout and the Bayesian networks. Dropout predictions are highly confident everywhere, except for at the decision boundaries between the classes. In contrast, the Bayesian networks exhibit high uncertainty in most areas, especially where little data is observed. When we increase the amount of data, we can see that the Bayesian networks gradually get more certain about predictions, and the entropies (as desired) start to converge towards the data-generative ones, while for dropout at a given rate, the uncertainties do not reduce.  It should be noted that there is no under-fitting happening, as we have close to 100$\%$ accuracy during training for all the methods. As a final observation, we see that the dense BNN typically has slightly less uncertainty than LBBNN with LRT and FLOW. Although, we can not say much about how good/bad this is, since it is difficult to obtain the true uncertainties by doing a reversible jump MCMC in the settings of LBBNN.   

\begin{figure}
    \centering
    \includegraphics[width=.47\textwidth]{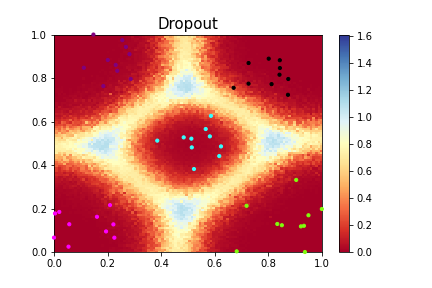}\hfill
    \includegraphics[width=.47\textwidth]{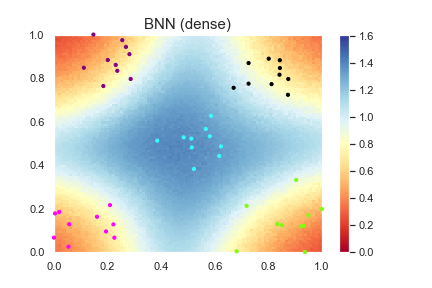}
    \vfill
    \includegraphics[width=.47\textwidth]{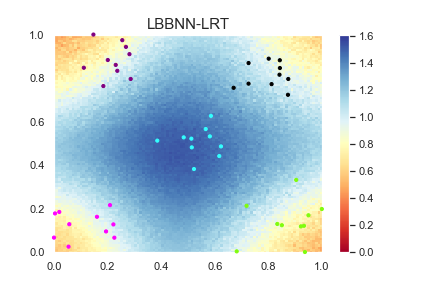}
    \hfill
    \includegraphics[width=.47\textwidth]{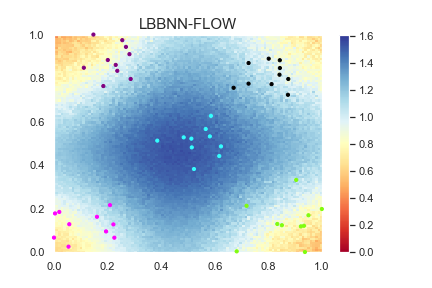}
    \caption{Entropy with 10 samples from each cluster}
    \label{figure:entropy50}
\end{figure}
\begin{figure}
    \centering
    \includegraphics[width=.47\textwidth]{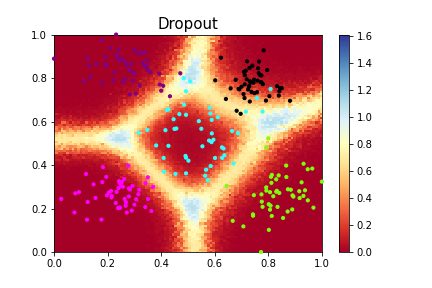}\hfill
    \includegraphics[width=.47\textwidth]{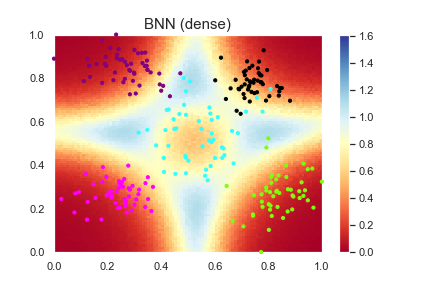}
    \vfill
    \includegraphics[width=.47\textwidth]{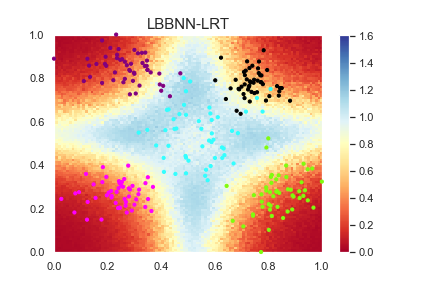}
    \hfill
    \includegraphics[width=.47\textwidth]{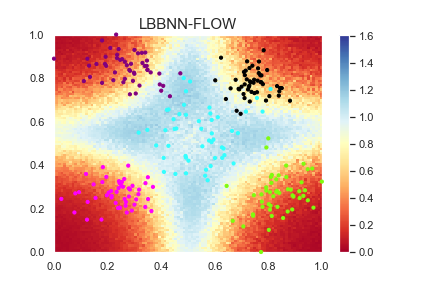}
    \caption{Entropy with 50 samples from each cluster}
    \label{figure:entropy250}
\end{figure}
\begin{figure}
    \centering
    \includegraphics[width=.47\textwidth]{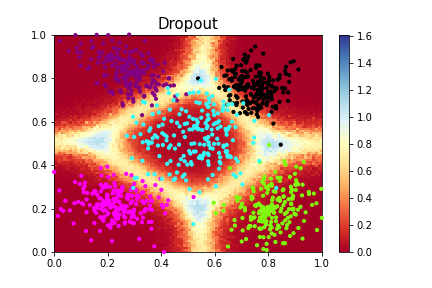}\hfill
    \includegraphics[width=.47\textwidth]{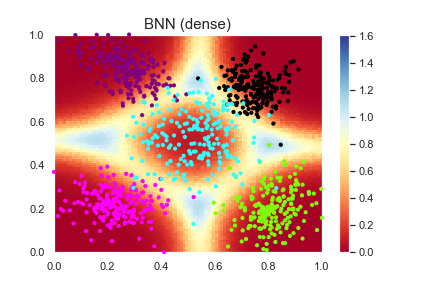}
    \vfill
    \includegraphics[width=.47\textwidth]{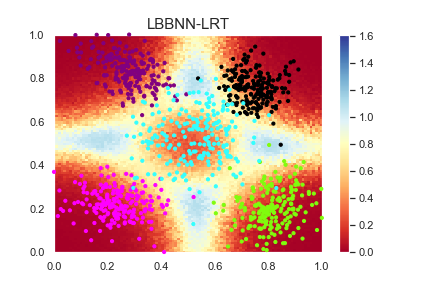}
    \hfill
    \includegraphics[width=.47\textwidth]{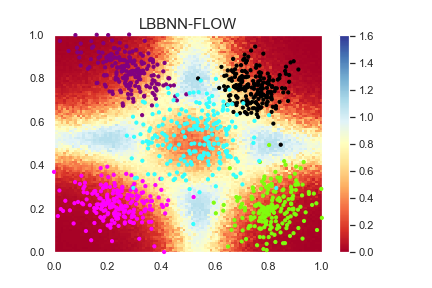}
    \caption{Entropy with 200 samples from each cluster}
    \label{figure:entropy1000}
\end{figure}
\begin{figure}
    \centering
    \includegraphics[width=.47\textwidth]{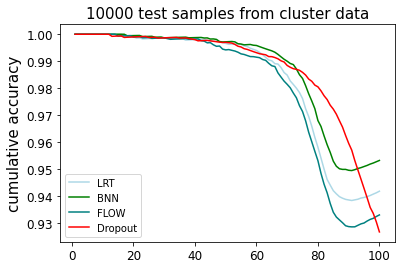}\hfill
    \includegraphics[width=.47\textwidth]{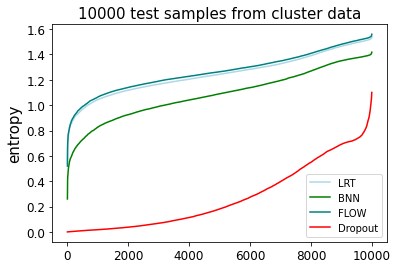}
    \vfill
    \includegraphics[width=.47\textwidth]{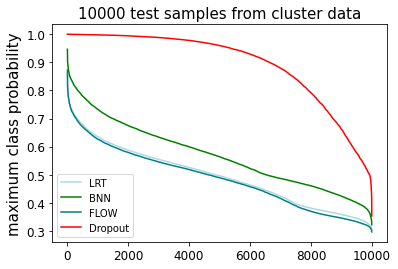}
    \caption{Top left, cumulative accuracy (100 samples at a time), where each point is the accuracy for the corresponding data points. Top right, entropy sorted from low to high. Bottom, maximum class probability sorted from high to low.}
    \label{figure:clusterstestdata}
\end{figure}

Additionally, we perform an experiment where we generate 10\,000 test samples (2\,000 from each cluster), after training with 50 samples (10 from each cluster). After training, we compute the entropy of the predictive distribution on the test data and sort the data from lowest to highest entropy. We also sort the samples based on the maximum class probability and compute the cumulative accuracy (with 100 data samples at a time). By that we mean that we start with the accuracy for the 100 most confident predictions, followed by 100 less confident predictions, and so on until we reach 100 of the least confident predictions. The results are in \hyperref[figure:clusterstestdata]{Figure 9}. With dropout, the maximum class probability is typically very high (i.e. we are extremely certain about which class the sample belongs to). After the first 5\,000 (sorted) samples, the output probability for the most likely class is at around $95\%$. With LRT and FLOW, on the other hand, it has dropped to roughly $50\%$. This mirrors what we saw earlier, dropout has high certainty most of the time. Despite this, we see that in this experiment the Bayesian methods have higher predictive accuracy than dropout for the cases with the most uncertainty. 
\begin{figure}[!h]
    \centering
    \includegraphics[width=.47\textwidth]
    {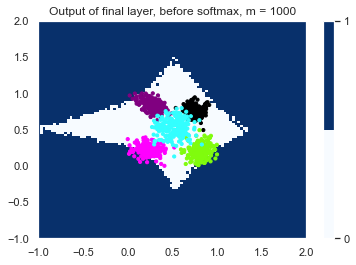}
    \caption{Out of distribution entropy, where dark blue corresponds to the OOD data detected by the BNN, and white is the in-distribution data.}
    \label{figure:oodentropy}
\end{figure}

As a final illustration, we consider an experiment where we take the maximum model averaged pre-activation output (pre-softmax) of the last layer (i.e. just before applying the softmax function) as a measure instead of using entropy. We use the training data ($m = 1000$) to generate an empirical confidence interval for the model-averaged pre-activation outputs for all the classes. We use a one-sided 95$\%$ confidence interval on the upper bound. During testing, we generate a sample over a grid, now between -1 and 2 in both dimensions, and take the highest model-averaged  pre-activation output. We then check whether it falls within the empirical confidence interval or not. The results are shown in \hyperref[figure:oodentropy]{Figure 10}. We see that in the regions with extremely low entropy, we can detect out-of-distribution data. This shows that using maximal entropy for out-of-distribution data as suggested in \cite{louizos2017multiplicative} might not be optimal. However, we still see the potential of BNNs to differentiate between in and out-of-domain uncertainty using the pre-activation values of the output of BNNs. We do not go any further here and leave this topic for future research.

\section{Discussion}
We have demonstrated that increasing the flexibility in the variational posterior distribution with normalizing flows improves the predictive power compared to the baseline method (with mean-field posterior) while obtaining more sparse networks, despite having a looser variational bound than the mean-field approach. Also, the flow method performed best on a variable selection problem, where the mean-field approaches struggle with highly correlated variables. More generally, we argue that Bayesian neural networks (BNNs) are much better at obtaining realistic predictive uncertainty estimates than their frequentist counterparts, as they have higher uncertainty when data is sparse. We do not observe a big difference in the uncertainty estimates obtained with dense BNN compared to our approaches. Unlike dense BNNs, our methods have the additional advantage of being able to perform variable selection. The downside is that LBBNNs have an extra parameter per weight, making them less computationally efficient than dense BNNs. Using normalizing flows is a further computational burden as we must also optimize over all the extra flow parameters. 

In this paper, we use the same prior for all the weights and inclusion indicators, although this is not necessary. A possible avenue of further research could be to vary the prior inclusion probabilities, to induce different sparsity structures. Currently, we are taking into account uncertainty in weights and parameters, given some neural network architecture. In the future, it may be of interest to see if it is also possible to incorporate uncertainty in the activation functions. By having connections between the layers, we could learn to skip all non-linear layers if a linear function is enough. A possible application is to do a genome-wide association study (GWAS), using our method. Combining LBBNNs and GWAS has been proposed by \cite{demetci2021multi}, however, this only uses the mean-field posterior. With our normalizing flow approach, we can easily model dependencies within each SNP set, in addition to dependencies between the different SNP sets. 

\section*{Supplementary material}

\noindent \textbf{GitHub:} The code used for the experiments can be found at
\url{https://github.com/LarsELund/Sparsifying-BNNs-with-LRT-and-NF}


\vskip 0.2in
\bibliography{sample.bib}

\newpage
\appendix
\section{Convolutional architectures}\label{appendix:A}
For convolutional layers, the variational distribution is defined to be:
\begin{equation}\label{eq:varpostconv}
\begin{split}
q_{\boldsymbol{\theta}}({\textbf{W}|\boldsymbol{\Gamma}},\boldsymbol{z})&=\prod\limits_{i=1}^{n_{h}}\prod\limits_{j=1}^{n_{w}}\prod\limits_{k=1}^{n_{f}}[\gamma_{ijk}\mathcal{N}(z_k\tilde \mu_{ijk},\tilde \sigma_{ijk}^{2}) + (1-\gamma_{ijk})\delta(w_{ijk})] \\
q_{\tilde \alpha_{ijk}}(\gamma_{ijk}) &= \text{Bernoulli}(\tilde \alpha_{ijk}),
\end{split}
\end{equation}
where $n_h$, $n_w$, and $n_f$ denote the height, width, and number of filters in the convolutional kernel. 

For the convolutional layers, we use the following for the inverse normalizing flows:
\begin{equation}
\begin{split}
    \boldsymbol{\nu} &= (\left(\text{Mat} (\boldsymbol{W}\odot\boldsymbol{\Gamma})\textbf{e})\otimes \textbf{d}_1\right)(\boldsymbol{1}\odot (n_{\text{h}}n_{\text{w}})^{-1})\\
     \log \boldsymbol{\tau^2}&=(\left(\text{Mat} (\boldsymbol{W}\odot\boldsymbol{\Gamma})\textbf{e})\otimes \textbf{d}_2\right)(\boldsymbol{1}\odot (n_{\text{h}}n_{\text{w}})^{-1}).
    \end{split}
\end{equation}
Here, $\text{Mat}(\cdot)$ denotes the matricisation operator (as defined in \citep{louizos2017multiplicative}), i.e. changing the shape of a multidimensional tensor into a matrix. 

\section{Data for predictive uncertainty experiments}\label{appendix:B}
For the predictive uncertainty experiment, we generate data from the following Gaussian distributions:
\begin{align*}G_1 &\sim \mathcal{N}\left(\begin{pmatrix} -8 \\ -8\end{pmatrix},\begin{pmatrix}
6 & -1\\
-1 & 3.5 
\end{pmatrix}
\right),\\
G_2&\sim \mathcal{N}\left(\begin{pmatrix} 6 \\ 6\end{pmatrix},\begin{pmatrix}
0 & 3\\
3 & 0 
\end{pmatrix}
\right),\\
G_3&\sim \mathcal{N}\left(\begin{pmatrix} -7 \\ 8\end{pmatrix},\begin{pmatrix}
-3 & 4\\
-5 & 1 
\end{pmatrix}
\right),\\
G_4&\sim \mathcal{N}\left(\begin{pmatrix} 8 \\ -8\end{pmatrix},\begin{pmatrix}
0 & 5\\
4 & 2 
\end{pmatrix}
\right),\\
G_5&\sim \mathcal{N}\left(\begin{pmatrix} 0 \\ 0\end{pmatrix},\begin{pmatrix}
0 & 9\\
9 & 0 
\end{pmatrix}
\right).
\end{align*}

\end{document}